\RequirePackage{amsmath}
\documentclass{llncs}

\usepackage{marvosym}
\usepackage{amsmath}
\usepackage{cases}
\usepackage{amssymb}

\DeclareMathOperator{\sign}{sgn}
\usepackage{lipsum}
\usepackage{graphicx}
\graphicspath{{resources/}}
\usepackage{subfigure}

\usepackage{hyperref}
\usepackage{fancyhdr}
\begin{document}
\title{Sparseout: Controlling Sparsity in Deep Networks}%
\author{Najeeb Khan \and
Ian Stavness
}

\institute{Dept. of Computer Science, University of Saskatchewan, Canada 
\email{najeeb.khan@usask.ca}\\ 
\email{ian.stavness@usask.ca}
}

\maketitle              

\begin{abstract}
Dropout is commonly used to help reduce overfitting in deep neural networks. Sparsity is a potentially important property of neural networks, but is not explicitly controlled by Dropout-based regularization. In this work, we propose Sparseout a simple and efficient variant of Dropout that can be used to control the sparsity of the activations in a neural network. We theoretically prove that Sparseout is equivalent to an $L_q$ penalty on the features of a generalized linear model and that Dropout is a special case of Sparseout for neural networks. We empirically demonstrate that Sparseout is computationally inexpensive and is able to control the desired level of sparsity in the activations. We evaluated Sparseout on image classification and language modelling tasks to see the effect of sparsity on these tasks. We found that sparsity of the activations is favorable for language modelling performance while image classification benefits from denser activations. Sparseout provides a way to investigate sparsity in state-of-the-art deep learning models. Source code for Sparseout could be found at \url{https://github.com/najeebkhan/sparseout}.
\end{abstract}

\section{Introduction}
\label{sec:intro}

Sparsity is often thought to be a desirable property for artificial neural networks. This is likely rooted in early neuroscience studies that discovered sparse coding in the visual cortex \cite{olshausen1997sparse} hypothesizing that at any given time, only a small number of neurons are used to encode sensory information. Sparsity has been observed both in connectivity \cite{morris2003anatomical} and representation \cite{olshausen1997sparse}. To mimic sparse coding from brain studies, researchers have devised approaches to encourage sparsity when training ANNs.

Sparsity has been used to regularize models by imposing a \emph{sparsity} constraint on the activations of the neural network \cite{thom2013sparse}. Many useful properties are ascribed to sparsity in the literature. It has been hypothesized that neurons that are rarely active are more interpretable than those that are active most of the time \cite{hinton2010practical}; images of natural world objects can be described in terms of sparse statistically independent events; neural networks with sparsity constraints learn filters that resemble the mammalian visual cortex area V1 \cite{olshausen1997sparse} and area V2 \cite{lee2008sparse}; and sparsity allows faster learning \cite{schweighofer2001unsupervised}. 

One of the main motivations behind sparsity based training methods is the biological plausibility of these algorithms. However, recent studies have questioned the pervasiveness of neural sparsity. The biological studies that provide evidence for sparsity are performed when the subject is passive and in reality sparsity might not be the mechanism that the brain uses in active tasks \cite{spanne2015questioning}. Empirically for DNNs, new methods that may discourage sparsity, such as Maxout \cite{goodfellow2013maxout} and DARC1 \cite{kawaguchi2017generalization}, have achieved better performance than sparse methods in certain domains such as computer vision. Therefore it is not clear whether or not sparsity is a generally desirable property for DNNs. We hypothesize that sparsity will benefit some learning tasks and hinder others. Therefore, new DNN training approaches that include the flexibility to either encourage sparsity, where necessary, and discourage sparsity otherwise could provide improved task performance.

There are many approaches for affecting the sparsity of a DNN during training including the use of certain activation functions such as rectified linear units \cite{glorot2011deep}. One of the main ways is through regularization and several deterministic regularization algorithms have been proposed to train deep neural networks with sparse weights \cite{hanson1989comparing,lecun1989optimal,han2015learning} and sparse activations \cite{chauvin1989back,mrazova2007improved,wan2009enhancing,glorot2011deep,liao2016learning}. 

Training deep neural networks with deterministic regularization and backpropagation results in correlated activities of the neurons. To prevent such co-adaptations as well as regularize the models, stochastic regularization methods are used. Stochastic methods such as Dropout, Bridgeout and Shakeout have been shown to be equivalent to ridge, bridge and elastic-net penalties on the model weights. Previous stochastic regularization methods that explicitly encourage sparsity, i.e., Shakeout and Bridgeout require a new set of masked weights per training example in a mini-batch making them computationally expensive. Therefore, these existing methods cannot be applied to large fully connected architectures. Likewise, Shakeout and Bridgeout cannot be easily applied to other convolutional architectures such as DenseNet and Wide-ResNet that provide current state-of-the-art performance for image classification, because they cannot be used with  highly optimized black-box implementations such as cuDNN \cite{chetlur2014cudnn}. 

In this paper, we propose Sparseout, a stochastic regularization method that is capable of either encouraging or discouraging sparsity in deep neural networks. It provides an $L_q$-norm penalty on the network's activations and therefore can vary activation sparsity by its $q$ parameter. The computational cost of Sparseout is comparable to Dropout and it can be applied to existing optimized CNN and LSTM blocks, making it applicable to state-of-the-art architectures. We provide theoretical and empirical results demonstrating the bridge-regularization capability of Sparseout. We use Sparseout to evaluate whether or not sparsity is beneficial for two distinct learning tasks: image classification and language modeling.

\section{Related work}
\label{sec:rw}

Due to the over-parameterization of deep neural networks, they suffer from large generalization error, specifically, when the dataset size is relatively small. This phenomenon is known as over-fitting. Generalization error is upper bounded by the model complexity \cite{shalev2014understanding} thus overfitting could be reduced by controlling the complexity of the model.

One way to control the complexity of a model is to impose constraints on the parameters of the model such as the weights in the neural networks. Such model regularization methods can be classified into deterministic and stochastic methods. Deterministic methods either remove redundant weights \cite{han2015learning} or penalize large magnitude weights. Weight penalties are imposed by adding a regularization term to the loss function consisting of a norm of the weight matrix \cite{neyshabur2015norm}. 

Stochastic methods randomly perturb the weights so as to achieve minimal co-dependency between neurons \cite{srivastava2014dropout,kang2016shakeout,khan2018bridgeout} as well as regularizing the model at the same time. Stochastic regularization has become the standard practice in training deep learning models and have outperformed deterministic regularization methods on many tasks. Stochastic regularization techniques have a Bayesian model averaging interpretation as well as they posses an equivalence to weight penalties for linear models. In terms of Bayesian estimation, weight penalties are equivalent to imposing a prior distribution on the model weights.

Beside the weight penalty interpretation, a reason for the effectiveness of stochastic regularization methods could be the prevention of correlated activations. It has been shown that high correlation between activations of the neurons results in overfitting. DeCov \cite{cogswell2015reducing}, reduces overfitting by adding a penalty term to the cost function consisting of the co-variances among the activations of the neurons over a mini-batch.

Another approach to control model complexity, inspired by sparse coding \cite{olshausen1997sparse}, is to impose a \emph{sparsity} constraint on the activations of the neural network \cite{thom2013sparse}. 
To encourage sparsity of the activations, an $L_1$ norm of the activations is added to the cost function \cite{lee2008sparse}. 
Another form of penalty is to add the KL-divergence of the expected activations and a preset target sparsity value $\rho$ \cite{hinton2010practical}. Liao et al. have used a clustering approach to obtain sparse representation by encouraging activations to form clusters \cite{liao2016learning}. 

Another related technique that normalizes activations in the network so as to have zero mean and unit variance is Batchnorm \cite{ioffe2015batch}. Although, the primary purpose of Batchnorm is accelerating training/optimization of the neural network rather than regularization, Batchnorm has reduced the need for stochastic regularization in certain domains. 
The above mentioned sparsity-inducing methods are deterministic and thus may result in correlated activations. In this paper we propose Sparseout that implicitly imposes an $L_q$ penalty on the activations thus allowing us to choose the level of sparsity in the activations as well as the stochasticity preventing correlated neural activities. Sparseout is different than Bridgeout \cite{khan2018bridgeout} in that it is applied to activations rather than the weights. Therefore, Sparseout is orders of magnitude faster and practical than Bridgeout. We believe that Sparseout is the first theoretically-motivated technique that is capable of simultaneously controlling sparsity in activations and reducing correlations between them, besides being equivalent to Dropout for $q=2$.

\section{Sparseout}
\label{sec:so}
Consider a feedforward neural network layer $l$, the output of $l$-th is given by 
\begin{equation}
    \boldsymbol{a}^{l} = \sigma \big( \boldsymbol{W}^{l}\boldsymbol{a}^{l-1} + \boldsymbol{b}^{l} \big),
\end{equation}
where $\boldsymbol{W}^{l}$ and $\boldsymbol{b}^{l}$ are the weight matrix and bias vector for the $l$-th layer, $\sigma$ is a non-linear activation function and $\boldsymbol{a}^{l-1}$ is the output of the previous layer.

The Sparseout perturbed output of the $l$-th layer $\boldsymbol{\widetilde{a}}^{l}$ is given by 
\begin{equation}
\boldsymbol{\widetilde{a}}_i^{l} = \begin{cases} 
      \boldsymbol{a}_i^{l} - |\boldsymbol{a}_i^{l}|^{q/2} & \boldsymbol{r}_i = 0 \\
      \boldsymbol{a}_i^{l} + |\boldsymbol{a}_i^{l}|^{q/2}\big(\frac{1-p}{p}\big) & \boldsymbol{r}_i =\frac{1}{p} \\
   \end{cases}
   \label{eq:so}
\end{equation}
where $\boldsymbol{r}$ is a random mask vector randomly sampled from a Bernoulli distribution with probability $p$ and scaled by $1/p$ and $q$ specifies the normed space to which the activations are restricted. Since the random mask is scaled by $1/p$ during training, no changes to the neural network are needed during testing. Since the training of the neural networks is performed using the back-propagated gradients of the error, the gradient of the Sparseout perturbed output is given by
\begin{equation}
    \frac{ \partial \boldsymbol{\widetilde{a}}_i^{l} }{\partial \boldsymbol{a}_i^{l}} = 1 + \frac{q}{2} |\boldsymbol{a}_i^{l}|^{\frac{q}{2}-1} \big(\boldsymbol{r}_i - 1\big) \sign \big(\boldsymbol{a}_i^{l}\big),
\end{equation}
where $\sign$ is the sign function.

Since Sparseout operates on the activations of the neural networks similar to Dropout, Sparseout can be implemented with minimal changes to the existing Dropout implementation. Sparseout can be used with the highly optimized black-box implementations such as cuDNN \cite{chetlur2014cudnn}. The above Sparseout formulation is applicable to any feedforward network layer such as convolutional or fully connected layers as well as layers in recurrent neural networks.

\begin{theorem}
	\label{thm:sparseout}
	Sparseout is equivalent to an $L_q$ penalty on the features of a generalized linear model.
\end{theorem}
\vspace{-0.2cm}
For a generalized linear model with parameters $\boldsymbol{\beta}$, log partition function $A$ and the perturbed design matrix $\boldsymbol{\widetilde{X}}$ of dimension $n \times d$, the negative log likelihood function could be split into a mean squared error term and a penalty term~\cite[eq.~6]{wager2013dropout} given by
\begin{equation}
R(\boldsymbol{\beta})   \approx  \sum_{i=1}^n \frac{ A''(\boldsymbol{X}_i \cdot\boldsymbol{\beta})}{2} Var[\widetilde{\boldsymbol{X}}_i \cdot\boldsymbol{\beta}] 
\label{eq:quad_penalty}
\end{equation}
For the Sparseout perturbation $\boldsymbol{\widetilde{X}}_{i,j} =\boldsymbol{X}_{i,j} [1 + |\boldsymbol{X}_{i,j} |^{\frac{q-2}{2}}(\boldsymbol{r}_j - 1)]$, the variance of $\widetilde{\boldsymbol{X}}_i \cdot\boldsymbol{\beta}$ is given by
\begin{equation}
Var[\widetilde{\boldsymbol{X}}_i \cdot\boldsymbol{\beta}] = \sum_{j=1}^d \frac{1-p}{p} |\boldsymbol{X}_{i,j}|^{q}\beta_j^2
\end{equation}
Substituting in Equation \ref{eq:quad_penalty} we have 
\begin{align}
\hat{R}(\boldsymbol{\beta}) & =  \frac{1-p}{2p} ||\Gamma\boldsymbol{X}||_q^q, 
\end{align}
where $\Gamma=[\boldsymbol{\beta}^TV(\boldsymbol{\beta})\boldsymbol{\beta}]^{\frac{2}{q}}$ and $V(\boldsymbol{\beta})=Diag(A)$. \hfill $\blacksquare$
\begin{theorem}
    \label{thm:do}
    For non-negative activation functions, Dropout is equivalent to Sparseout when $q=2$.
\end{theorem}
\vspace{-0.2cm}
Setting $q=2$ in Equation \ref{eq:so} and considering the fact that $\boldsymbol{a}$ is non-negative, we have $\boldsymbol{\widetilde{a}}_i^{l} = \boldsymbol{r}_i \boldsymbol{a}_i^{l}$, which is identical to the Dropout perturbation \cite{srivastava2014dropout}.  \hfill $\blacksquare$

$L_q$-normed spaces with different values of $L_q$ exhibit different sparsity charactersitics. For $q<2$ the norm space is sparse while for $q>2$ the norm space is dense \cite{park2011bridge}. With Sparseout we can select the norm space of the activations by choosing the value of the hyper-parameter $q$. Thus, Sparseout allows us to control the level of sparsity in the activations of the neural networks.

\section{Experimental results}
\label{sec:exp}

\subsection{Sparsity characterization}
\begin{figure}
    \centering
    \includegraphics[width=\linewidth]{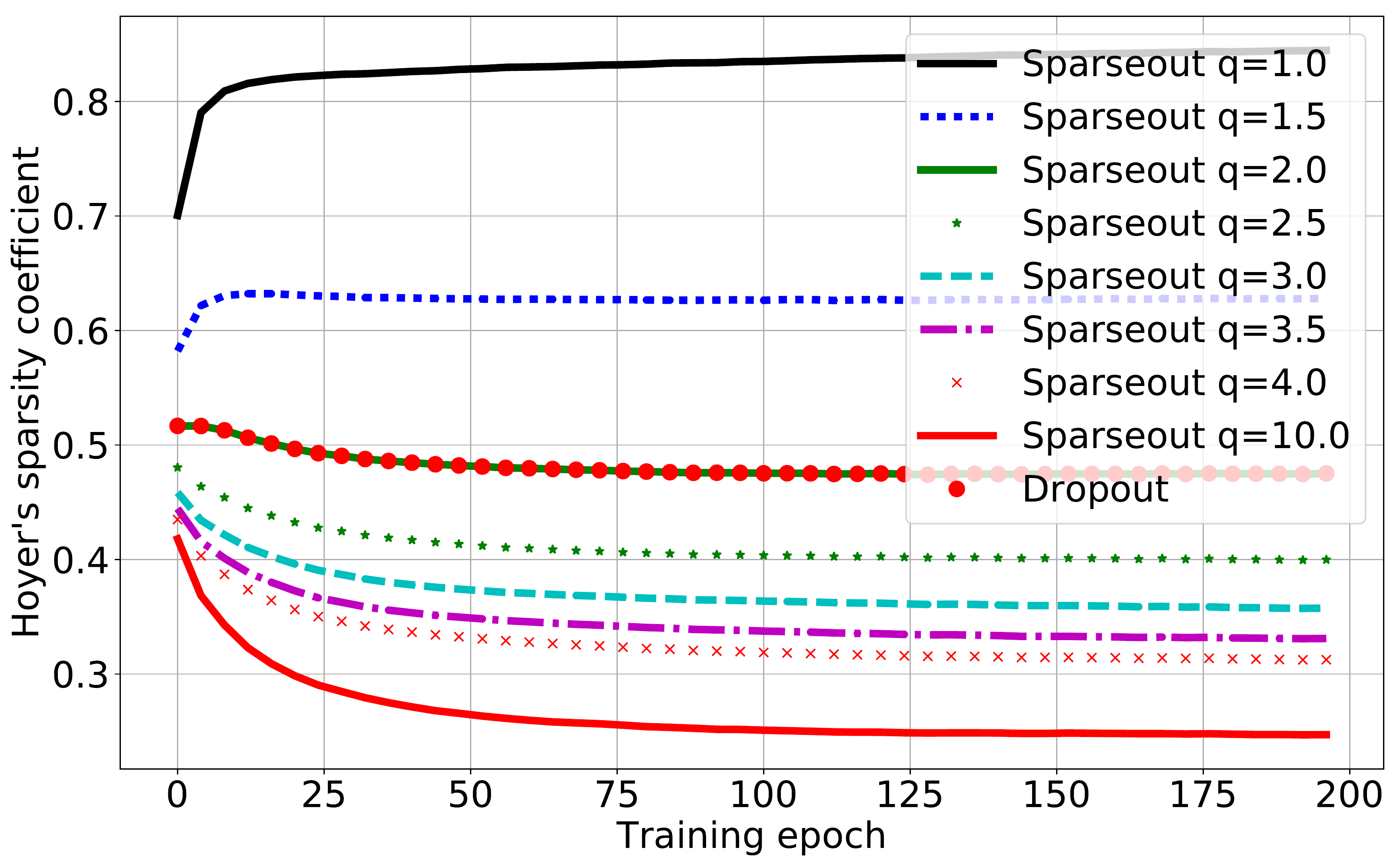}
    \caption{Average Hoyer's sparsity measure of the hidden layer activations calculated over the MNIST test set for an autoencoder trained on MNIST with Dropout versus Sparseout with different values of $q$.
    }
    \label{fig:ae_sparsity}
\end{figure}

To  verify that Sparseout is capable of controlling sparsity of a neural network's activations, we train an autoencoder with a hidden layer of $512$ rectified linear units on the MNIST dataset. Dropout and Sparseout are applied to the hidden layer activations with $p=0.5$ and different values of $q$ for Sparseout. We measure sparsity of the hidden layer activations during testing (when no perturbations are applied to the activations). To measure sparsity we use the Hoyer's measure $H$ \cite{hoyer2004non}: 
\begin{equation}
    H(\mathbf{x}) = \frac{\sqrt{d} - \frac{|\mathbf{x}|_1}{|\mathbf{x}|_2}}{\sqrt{d} - 1}
\end{equation}
where $\mathbf{x}$ is a $d$-dimensional vector, $|\mathbf{x}|_1$ is the $L_1$-norm and  $|\mathbf{x}|_2$ is the $L_2$-norm of $\mathbf{x}$.
A vector consisting of equal non-zero values has $H=0$ while vectors only having one non-zero element has $H=1$. Figure~\ref{fig:ae_sparsity} shows the Hoyer's sparsity measure on the test set as the training progresses for different values of $q$. As the value of $q$ decreases below $2$, we see an increase in the sparsity of the activations, whereas for $q$ values greater than $2$ the sparsity is reduced. For $q=2$, Sparseout results in the same sparsity as Dropout. These results confirm our theoretical analysis that Sparseout can be used to control sparsity of the activations in the neural networks.

\subsection{Computational cost}
Sparseout is computationally efficient and incurs similar training cost as Dropout. We train an autoencoder with two hidden layer sizes on MNIST with a batch size of $128$ both on Nvidia GTX 1080 GPU. As shown in Table \ref{tab:computeCost}, Sparseout is only fractionally more expansive than Dropout while Bridgeout is an order of magnitude more expensive even for this simple model. Also doubling the hidden layer size results in a doubling of the execution time for Brigdeout while Sparseout and Dropout have almost constant execution time due to utilization of GPU parallelism.
\begin{table}[ht]
    \centering
    \caption{Average execution time in seconds per epoch for different types of stochastic regularization for an autoencoder with different hidden layer sizes.}
    \begin{tabular}{|c|c|c| c |c |} 
         \hline
         Hidden layer size & Backprop & Dropout & Sparseout & Bridgeout \\ [0.5ex]
         \hline
         1024 & 5.2 & 5.3 & 5.8 & 31.6 \\
         \hline
         2048 & 5.6 & 5.6 & 6.0 & 57.2 \\
         \hline
    \end{tabular}
    \label{tab:computeCost}
\end{table}
\subsection{Image classification}
Image classification is one of the key areas where deep neural networks have been highly successful achieving state-of-the-art results. 
%
The CIFAR datasets \cite{krizhevsky2009learning} are a standard benchmark for image classification. The CIFAR-$10$ dataset consists of color images of size $32 \times 32$ each belonging to one of the ten classes of objects. The dataset is divided into a training set of $50000$ images and a test set of $10000$ images. The CIFAR-$100$ dataset is similar to CIFAR-$10$ except that the  images are divided into $100$ classes of objects, thus making the classification task more harder than CIFAR-$10$. We used the standard pre-processing of mean and standard deviation normalization. Random cropping and random horizontal flips were used for data augmentation.

We use the wide residual network (WRN) architecture to evaluate the effect of sparsity on classification accuracy using Sparseout. WRNs achieved state-of-the-art accuracy on several image classification tasks including CIFAR-$10$ and CIFAR-$100$. WRNs are based on deep residual networks \cite{he2016deep} that use identity links between the input and output of each layer known as the residual connections, but they employ fewer and wider layers. The residual connections helps in training very deep neural networks consisting of upto a thousand layers.

\begin{figure}
    \centering
    \includegraphics[height=5cm]{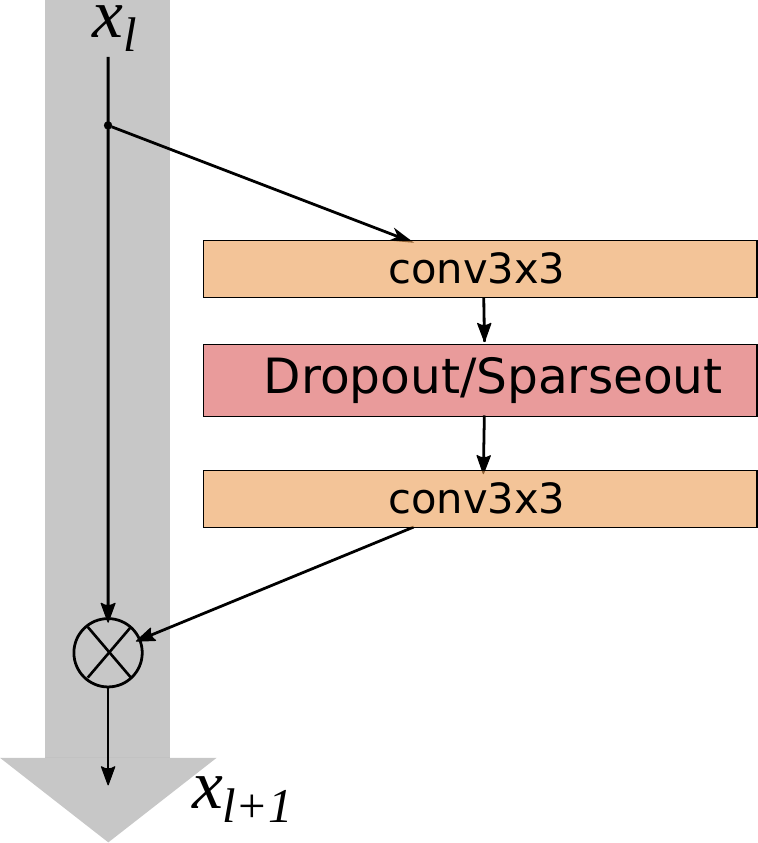}
    \caption{The basic building block of wide residual network architecture with either Dropout or Sparseout stochastic regularization.  Figure~adapted from \cite{zagoruyko2016wide}}
    \label{fig:wrn-ds}
\end{figure}

We employ a WRN with the basic building block shown in Figure~\ref{fig:wrn-ds}. The stochastic regularization is applied between the convolutional layers. Each convolutional layer is preceded by batch normalization and rectified linear unit activation function. In our experiments we use the WRN architecture WRN-28-10 with depth 28 and a widening factor of 10. A Dropout probability of $p=0.3$ was used. Stochastic gradient descent with a mini-batch of 64 was used to train the networks. The learning rate was annealed from $0.1$ at $60$, $120$ and $160$ epochs by a factor of $0.2$ as in the original WRN paper \cite{zagoruyko2016wide}. 

\subsubsection{Image Classification Results}
\begin{figure}
  \centering
  \subfigure[CIFAR-10]{\label{fig:wrn-cifar10}\includegraphics[width=0.49\textwidth]{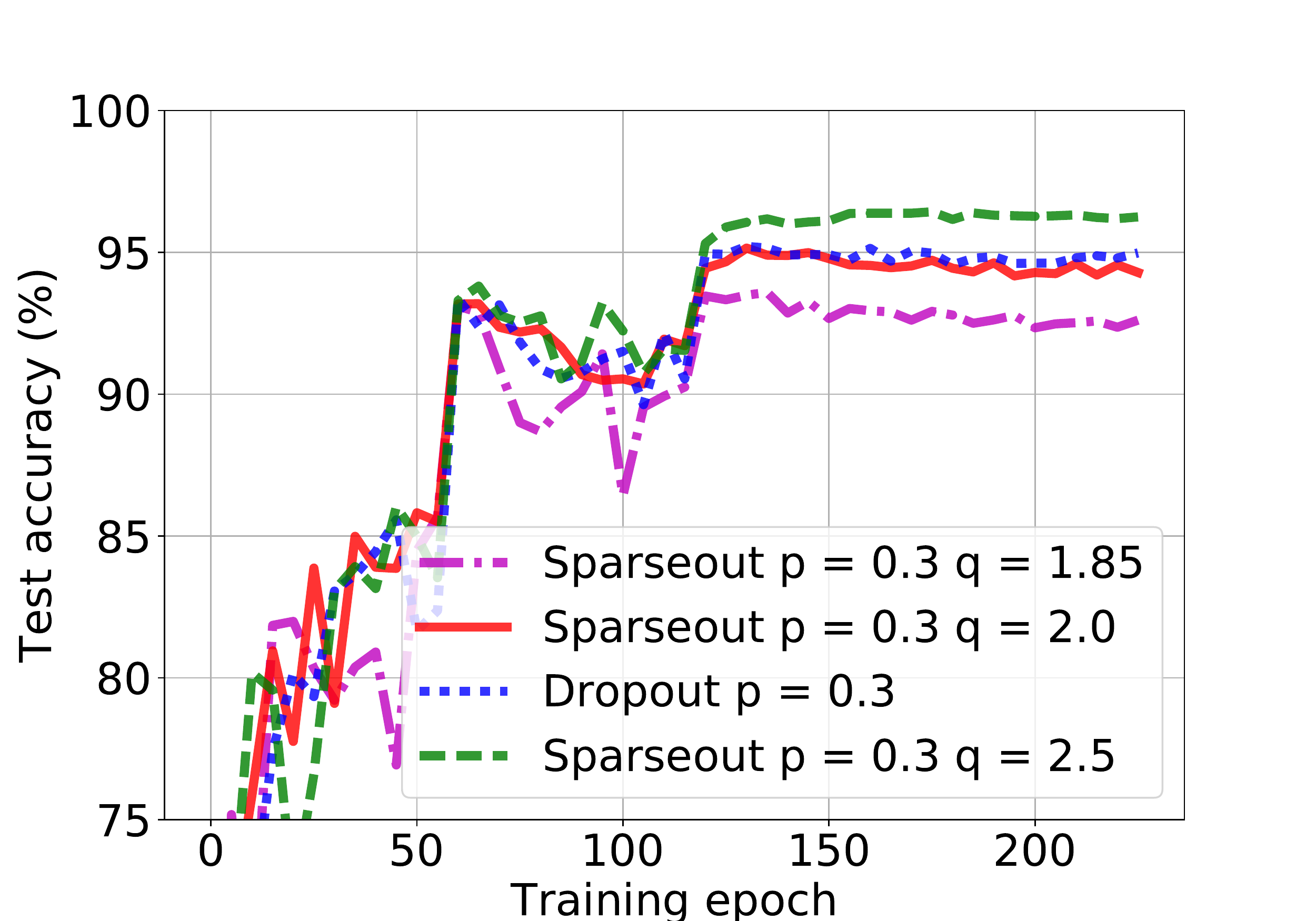}}
  \subfigure[CIFAR-100]{\label{fig:wrn-cifar100}\includegraphics[width=0.49\textwidth]{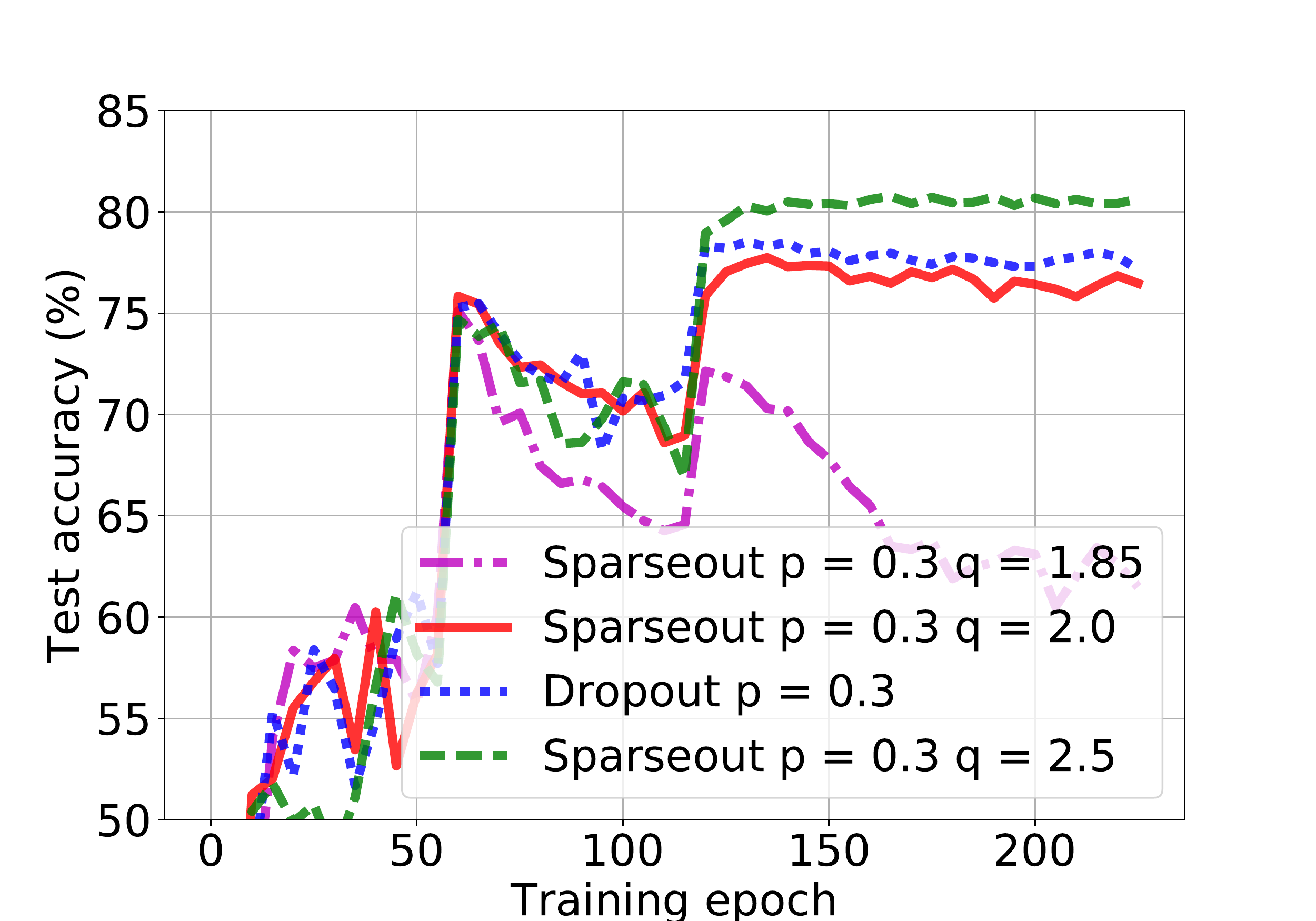}}
  \vspace{-0.2cm}
  \caption{Test accuracy during training of WRN on CIFAR-10 and CIFAR-100 for Dropout versus Sparseout with different values of $q$.}
  \label{fig:wrn-cifar}
\end{figure}

For image classification we found that Sparseout with $q>2$ resulted in better performance compared to values of $q<2$ as shown in Figure~\ref{fig:wrn-cifar}. For $q<2$ the accuracy drops as the training progresses beyond around $100$ epochs indicating overfitting. As shown in Table \ref{tab:cifar}, error rate for $q=2.5$ is about $1$ percent lower than Dropout for CIFAR-10 and $2.5$ percent lower for CIFAR-100. Our baseline results are comparable to the baselines reported in the literature for CIFAR-100 and better for CIFAR-10~\cite{zagoruyko2016wide,louizos2018learning}. 

\begin{table}[ht]
    \centering
    \caption{Mean test errors for the same WRN network trained with Dropout and Sparseout with different values of $q$. Lower sparsity (high $q$) results in lower test error.}%
    \begin{tabular}{|c|c| c |c |c|} 
         \hline
         Model & $p$ & $q$ & CIFAR10 & CIFAR100 \\ [0.5ex]
         \hline
         WRN-28-10-Sparseout & 0.3 & 1.5 & 7.58 & 25.87 \\
         \hline
         WRN-28-10-Sparseout & 0.3 & 1.85 & 5.92 & 24.65 \\
         \hline
         WRN-28-10-Sparseout & 0.3 & 2.0 & 4.72 & 21.91 \\
         \hline
         WRN-28-10-Dropout & 0.3 & - & 4.59  & 21.66 \\
         \hline
         WRN-28-10-Sparseout &0.3 & 2.5 & 3.63 & 19.07  \\
         \hline\hline
         WRN-28-10-Dropout \cite{zagoruyko2016wide} & \multicolumn{2}{|c|}{$p=$0.3} &  3.89 & 18.85\\ 
         \hline
         WRN-28-10-$L_{0_{hc}}$ \cite{louizos2018learning} & \multicolumn{2}{|c|}{-} & 3.83 & 18.75 \\
         \hline
    \end{tabular}
    \label{tab:cifar}
\end{table}

\subsection{Language Modelling}
Another task for which deep learning has been widely used is natural language processing (NLP). The dimensionalty of NLP tasks is very high and sparse; therefore, sparsity is likely to play an important role in such tasks. Language modelling (LM) assigns a probability to a sequence of words. LM is an important component of several NLP tasks such as speech recognition, information retrieval and machine translation among others. Since LM is a sequential task recurrent neural networks are used for it. Vanilla RNN are difficult to train due to vanishing and exploding gradients problem. To overcome these limitations, long short term memory (LSTM) models are used instead \cite{sundermeyer2012lstm}. 

The LSTM model is a type of recurrent neural network with layers consisting of \textit{memory cells}. The weights of the nodes in a memory cell learn the long term information while a node with a self-connected edge retains short term information. The input gate, forgetting gate and output gate help in controling the flow of information in the LSTM. For a detailed review of the LSTM formulation see Lipton et al. \cite{lipton2015critical}. 

We adapt the baseline LSTM architecture for word level language modelling from Merity et al.  \cite{merity2017regularizing}.  The model consists of 3 layers of 1150 units. To train the baseline model we used the same hyper-parameters used by Merity et al. \footnote{\url{https://github.com/salesforce/awd-lstm-lm}} except that we used only stochastic gradient descent for training. 

We replace variants of Dropout with variants of Sparseout in the LSTM model. Variational Dropout \cite{gal2015dropout} is replaced with variational Sparseout where a single random mask is used within a forward and backward pass. Embedding Dropout applied to the word embedding layer is similarly replaced with embedding Sparseout.

We evaluate the model on two standard word-level language modelling datasets where the task is to predict the next word and the performance is evaluated on perplexity which is the negative log likelihood raised to the exponent. The first dataset is the  Penn Treebank dataset \cite{Marcus1993} that contains $1$ million words and a vocabulary size of $10,000$. The second dataset is the WikiText-2 dataset \cite{merity2016pointer} which contains over $100$ million words and a vocabulary of size $30,000$. 

\subsubsection{Language modeling results}
\begin{figure}
  \centering
  \subfigure[PTB]{\label{fig:lstm-ptb}\includegraphics[width=0.49\textwidth]{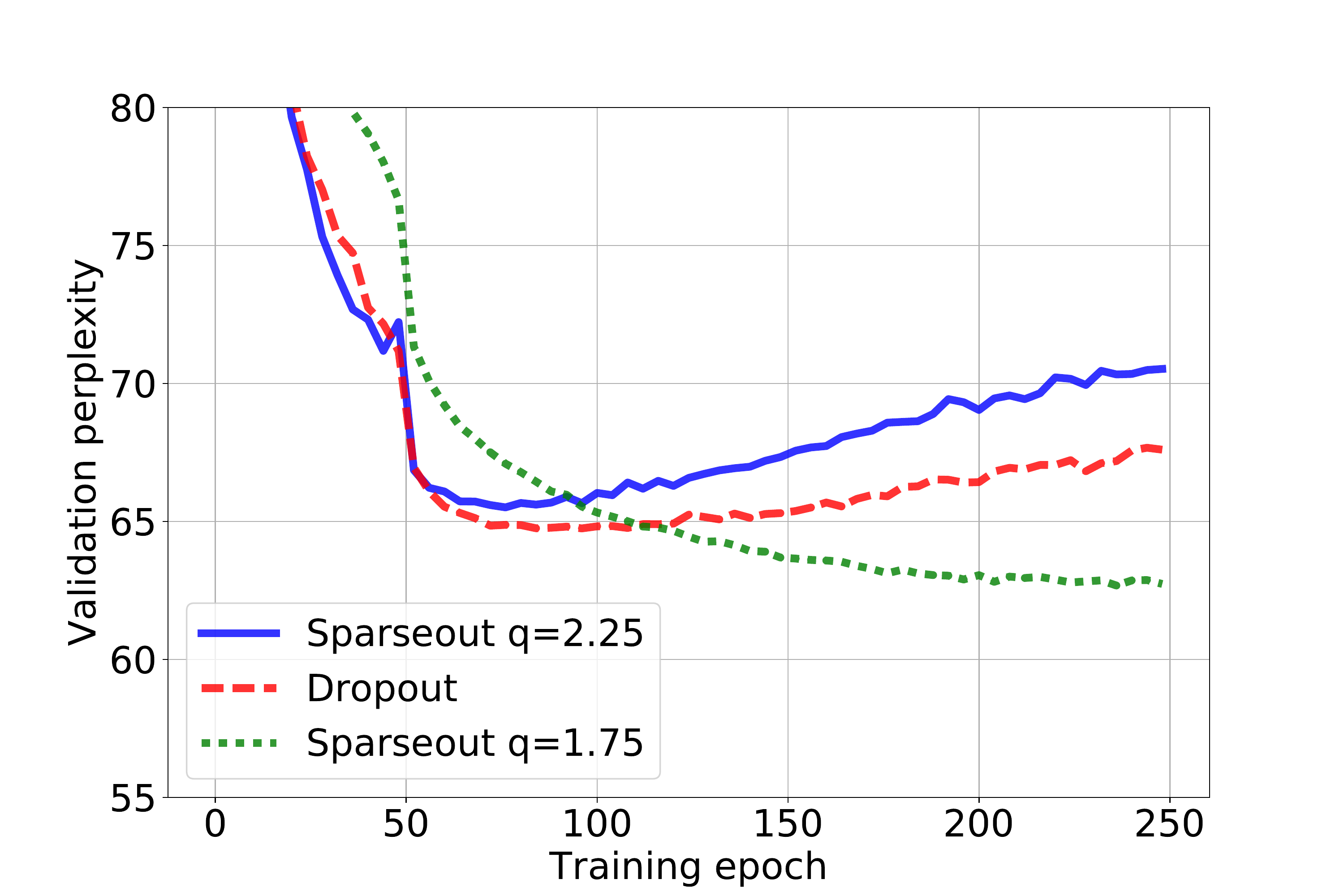}}
  \subfigure[Wiki-2]{\label{fig:lstm-wiki}\includegraphics[width=0.49\textwidth]{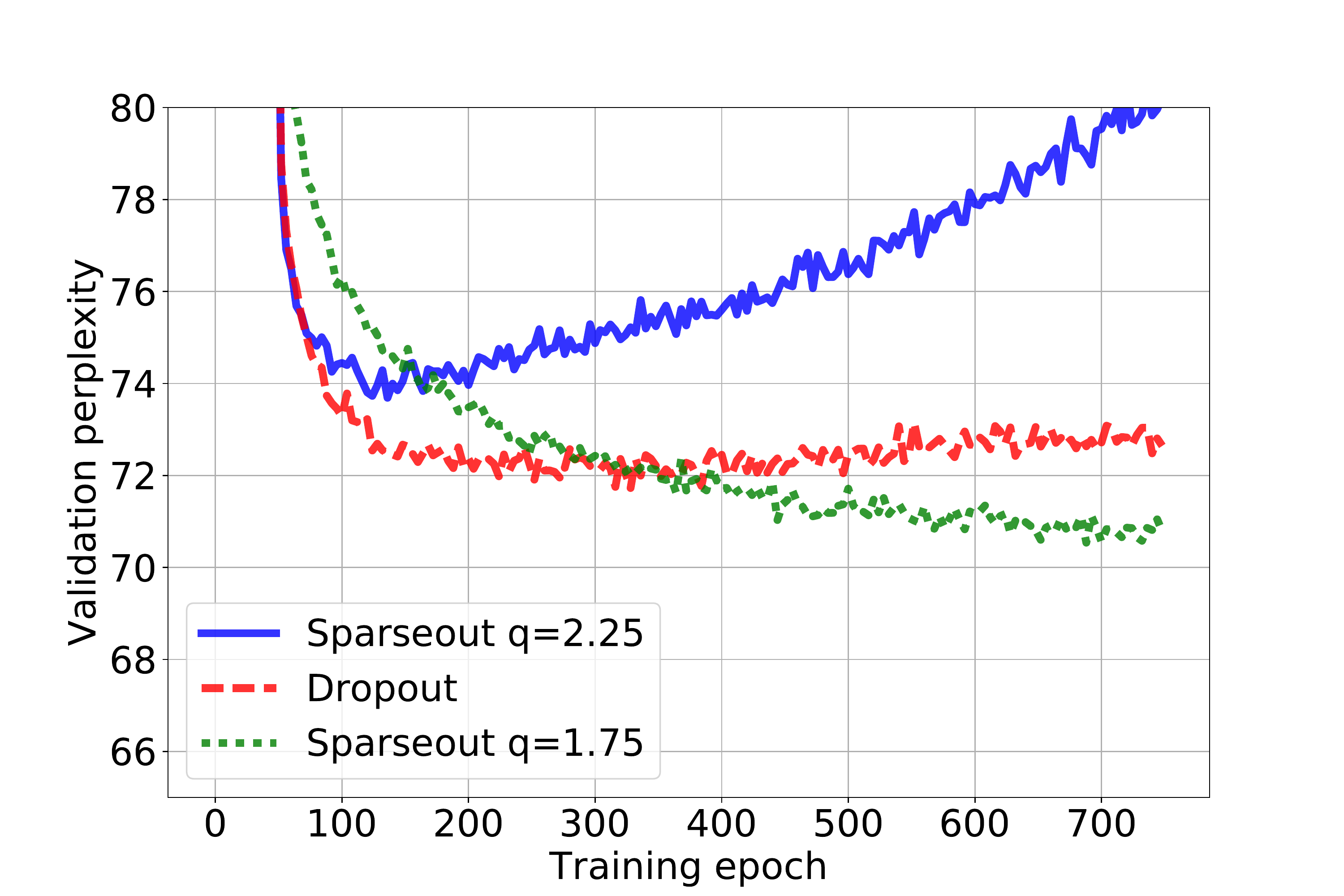}}
  \caption{Validation perplexity for language modeling on PTB and Wiki-2 datasets for LSTM models trained with Dropout, Sparseout with $q=2.25$ to reduce sparsity, and Sparseout with $q=1.75$ to increase sparsity.}
  \label{fig:lstm}
\end{figure}

Applying Sparseout with $q>2$ resulted in significant overfitting as shown in Figure \ref{fig:lstm}. For $q<2$ we found that Sparseout resulted in better prediction performance than Dropout. For PTB dataset Sparseout results in $2.5$ percent reduction in relative perplexity. For Wiki-2 dataset the reduction in relative perplexity is $1.25$ percent as shown in Table \ref{tab:lstm}.

\begin{table}[ht]
    \centering
    \caption{Single model test perplexities on PTB and Wiki-2 datasets for LSTM models trained with Dropout, Sparseout with $q=2.25$ to reduce sparsity, and Sparseout with $q=1.75$ to increase sparsity. Higher sparsity results in lower (better) relative perplexity.}
    \begin{tabular}{|c|c| c |c |c|} 
         \hline
         Model & Penn Tree Bank & WikiText-2 \\ [0.5ex]
         \hline
         LSTM-Sparseout ($q=2.25$)  & 62.7 & 70.18 \\
         \hline
         LSTM-Dropout  & 62.13 & 68.34 \\
         \hline
         LSTM-Sparseout ($q=1.75$)  & 60.57 & 67.17 \\
         \hline\hline
         AWD-LSTM-Dropout \cite{merity2017regularizing} &  57.3 & 65.8\\ 
         \hline
    \end{tabular}
    \label{tab:lstm}
\end{table}

\section{Discussion}
Existing literature is contradictory on whether sparsity is a good~\cite{chauvin1989back,mrazova2007improved,wan2009enhancing,glorot2011deep,liao2016learning,louizos2018learning} or bad~\cite{spanne2015questioning,rigamonti2011sparse,kawaguchi2017generalization,goodfellow2013maxout,gulcehre2014learned} property for deep neural networks. No previous study has evaluated sparse vs. non-sparse networks in a controlled fashion with stochastic regularization. In this study, we propose a new bridge-regularization scheme, Sparseout, which has the flexibility to control sparsity and the efficiency to be applied to large networks. 

We evaluated Sparseout with two distinct network architectures and machine learning tasks: CNNs for image classification and LSTMs for language modeling. 
Our empirical results show that lower sparsity improves image classification performance, whereas higher sparsity improves performance on language modeling. These results align with the fundamental differences between data types: relatively tiny densely-featured images vs. sparsely-featured high-dimensional language data. 

In this study, we chose the most suitable architecture for each task: CNNs for IID image classification and RNNs for sequential language modelling. Therefore, we evaluated task-architecture in a coupled manner.
 For each task, image classification or language modelling, we tested two datasets (CIFAR10/CIFAR100 or PTB/WikiText-2) and obtained consistent results regarding the benefit or lack thereof of sparse activations. It is possible, however, that the inherent sparse nature of convolutional layers requires spreading of the activations over all the neurons while enforced parsimony of representation is helpful for the fully connected gates in an LSTM. Therefore, decoupling the effect of data type from that of architecture is an important consideration we plan to investigate as future work.

\section*{Acknowledgments}
\label{sec:Acknowledgments}
This work was supported by the Natural Sciences and Engineering Research Council of Canada (NSERC).
\bibliographystyle{splncs}       
\bibliography{sparseout}   

\end{document}